%% file: example_paper.tex
\documentclass[10pt,twocolumn,letterpaper]{article}

\usepackage{cvpr}

\usepackage{microtype}
\usepackage{mathtools}
\usepackage{multirow}
\usepackage{cuted}
\usepackage{needspace}
\usepackage{tikz}
\usepackage{algorithm}
\usepackage{algorithmic}
\floatname{algorithm}{Pseudocode}

\definecolor{cvprblue}{rgb}{0.21,0.49,0.74}
\usepackage[breaklinks,colorlinks,allcolors=cvprblue]{hyperref}

\usepackage{amsmath}
\usepackage{amssymb}
\usepackage{amsthm}

\usepackage[capitalize,noabbrev]{cleveref}

\theoremstyle{plain}
\newtheorem{theorem}{Theorem}[section]
\newtheorem{proposition}[theorem]{Proposition}
\newtheorem{lemma}[theorem]{Lemma}

\theoremstyle{definition}
\newtheorem{definition}[theorem]{Definition}
\newtheorem{assumption}[theorem]{Assumption}
\theoremstyle{remark}

\usepackage{paralist}

\def\confName{CVPR}
\def\confYear{2026}

\title{Can We Change ``Stroke Size'' to Improve Flow Matching?}

\begin{document}

\author{
Yunwei Bai\thanks{Corresponding author.} \quad Ying Kiat Tan \quad Tsuhan Chen\\
National University of Singapore\\
{\small baiyunwei@u.nus.edu, yingkiat@u.nus.edu, dprtchen@nus.edu.sg}
}
\maketitle

\begin{strip}
    \centering
    \includegraphics[width=\textwidth]{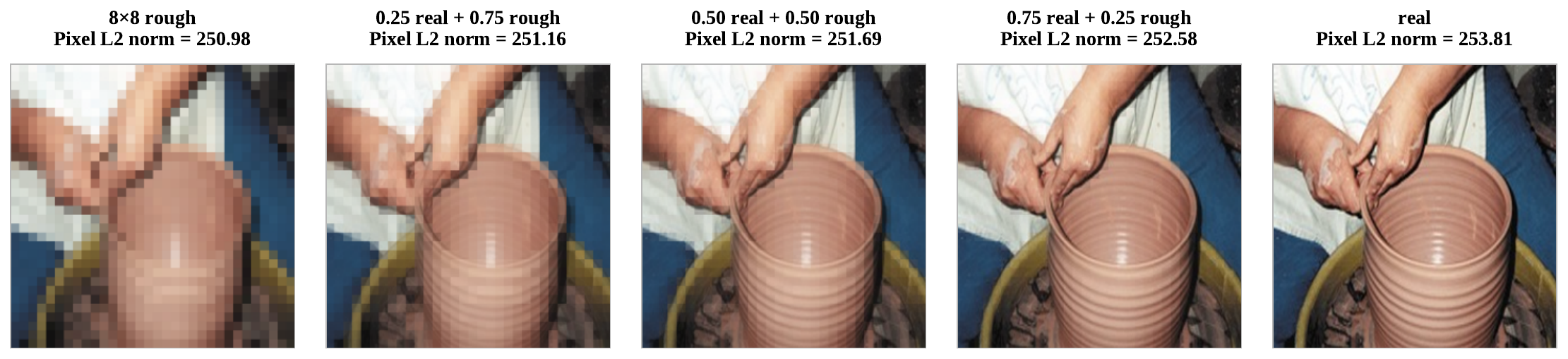}
    \captionof{figure}{An $8\times8$ roughening example and interpolants at
    $256\times256$ resolution, with their RGB pixel $\ell_2$ norms. The
    roughened endpoint has a lower norm and complexity than the original image.}
    \label{fig:latent-norm-8x8}
\end{strip}

\input{sections/1_intro}
\input{sections/2_related}
\input{sections/3_background_hypothesis}
\input{sections/4_method}
\input{sections/4_theory}
\input{sections/6_exp}

\input{sections/7_conclusion}

\input{sections/7_impact}
\onecolumn
\newpage
\twocolumn
{
\small
\bibliographystyle{ieeenat_fullname}
\bibliography{example_paper}
}

\onecolumn
\newpage
\twocolumn
\section*{LLM Usage Statement}
OpenAI Codex was used to assist with code inspection, experiment execution,
and manuscript editing. The authors verified all resulting analyses, claims,
and reported results.

\clearpage
\appendix
\onecolumn
\input{sections/8_app}

\end{document}

%% file: sections/1_intro.tex
\begin{abstract}
Deep-learning-based image generation models are useful but difficult to train.
Inspired by the varying
stroke sizes used in human painting, we hypothesize that applying ``rough strokes''
can attenuate prediction biases and improve flow-matching image generation.
We develop MultiStroke, which progressively varies the spatial granularity of
the clean-signal component in the velocity regression
target. A lightweight
corrector aligns adjacent roughness levels during sampling. On ImageNet-1K,
MultiStroke reduces FID by $22.5\%$ relative to a matched flow-matching
baseline, despite the limited corrector capacity.
Code is available at
\url{https://github.com/DaChengRuoQue/MultiStroke-Code}. A quick video
explanation is available at
\url{https://dachengruoque.github.io/MultiStroke-Web/}.
\end{abstract}

\section{Introduction}
\label{sec:intro}
Deep-learning-based image generation underlies widely used visual-content
creation systems \citep{saharia2022imagen,ramesh2022unclip}. Flow matching
learns a velocity field along a path between data and noise
\citep{lipman2023flow}. Predicting this field across the full path remains a
per-pixel regression problem. 
The pixel-scale prediction retains high-frequency targets even near the noisy end
of the path, while spectral discrepancies remain common in generated images
\citep{dzanic2019fourier}. 
Such a per-pixel target may be problematic in multiple ways.
For example, the mean squared error (MSE)
objective penalizes
errors uniformly across incorrect pixel predictions;
a predicted pixel that 
appears in proximity to 
the target pixel receives the same penalty magnitude as a 
predicted pixel that 
stays far away from the intended target. This is akin to using 
a uniform stroke throughout to replicate the target exactly, 
which may introduce unnecessary computational overhead alongside
sampling artifacts.

Prior work has addressed related optimization and generation difficulties by
moving the generative dynamics into compressed latent spaces
\citep{rombach2022ldm,vahdat2021lsgm}, replacing Gaussian corruption with
heat dissipation, blurring, or more general image transformations
\citep{rissanen2023heat,hoogeboom2023blurring,daras2023softdiffusion,
bansal2023cold}, and decomposing synthesis across resolutions or stages
\citep{ho2022cascaded,gu2023fdm,hoogeboom2023simple}. These approaches
primarily target computational efficiency, alternative generative paths, or
hierarchical high-resolution synthesis; they do not isolate the effect of
complexity reduction in the supervised velocity target over an existing 
flow-matching paradigm. To our knowledge,
MultiStroke is the \textbf{first}
work to apply 
down-up-sampling stroke size control (i.e., demonstrated by Figure 
\ref{fig:latent-norm-8x8}) as a complexity uppper bound reduction
mechanism, treating flow matching as a regression problem.
Note that we do \textbf{not} claim novelty on the coarse-to-fine 
control, which is not our focus; we apply the granularity 
stroke control just to ensure the output final image is clean. 

In essence, MultiStroke encourages 
the model to learn the coarse structure where possible, and then transition 
between different levels of roughness during sampling, via a small  
residual corrector. 
The roughening operator uses lightweight
down-up sampling and linear interpolation to form the  
coarse-to-fine sequence in pixel space, and the interpolation weight aligns  
with the noise weight; during sampling, the noise is undone 
alongside the roughness.
Since the coarse interpolation input and target 
are of lower complexity (e.g., defined in Section \ref{sec:background-hypothesis}), 
there can be 
complexity reduction benefits
such as high-frequency prediction artifact smoothing.
Besides, the roughness transition is small, bounded, and
constant in velocity; the down-up sampling interpolation 
is invertible, deterministic, and geometrically straight in the pixel-space
velocity field. Intuitively, this step is similar to confirming the exact 
location of the fine details given the local boundaries within which the
finer-detail pixels must remain.

Our contributions are:

\begin{compactenum}
    \item We hypothesize that reducing the data complexity provides
    a regularization benefit for flow matching, and propose a
    stroke-size control algorithm that modifies the roughness of
    supervision targets. A lightweight corrector partially corrects 
    roughness transition errors during sampling.
    \item We evaluate MultiStroke against a matched vanilla flow matching
    baseline trained on ImageNet-1K, achieving up to 22.5\% Fr\'echet Inception Distance (FID) improvement
    despite the corrector's limited capacity. We also conduct various mechanism studies.
\end{compactenum}

%% file: sections/2_related.tex
\section{Related Work}
\label{sec:related}

MultiStroke builds on diffusion, score-based generation, and flow matching
\citep{ho2020ddpm,song2021score,lipman2023flow}, including their latent-space
variants \citep{rombach2022ldm,vahdat2021lsgm}. Its closest connections are
structured corruption processes that progressively suppress high-frequency
content \citep{rissanen2023heat,hoogeboom2023blurring,bansal2023cold} and
coarse-to-fine generative methods based on resolution hierarchies
\citep{ho2022cascaded,gu2023fdm,hoogeboom2023simple}. Unlike these approaches,
MultiStroke retains one image resolution and one velocity model while varying
the spatial granularity of the flow-matching endpoint along the path. This
design also relates to spectral learning bias and multiscale regularization
\citep{rahaman2019spectral,ronneberger2015unet}. Appendix~\ref{app:related}
provides an extended discussion.

%% file: sections/3_background_hypothesis.tex
\section{Background and Hypothesis}
\label{sec:background-hypothesis}
\begin{figure*}[ht]
    \centering
    \includegraphics[width=\textwidth]{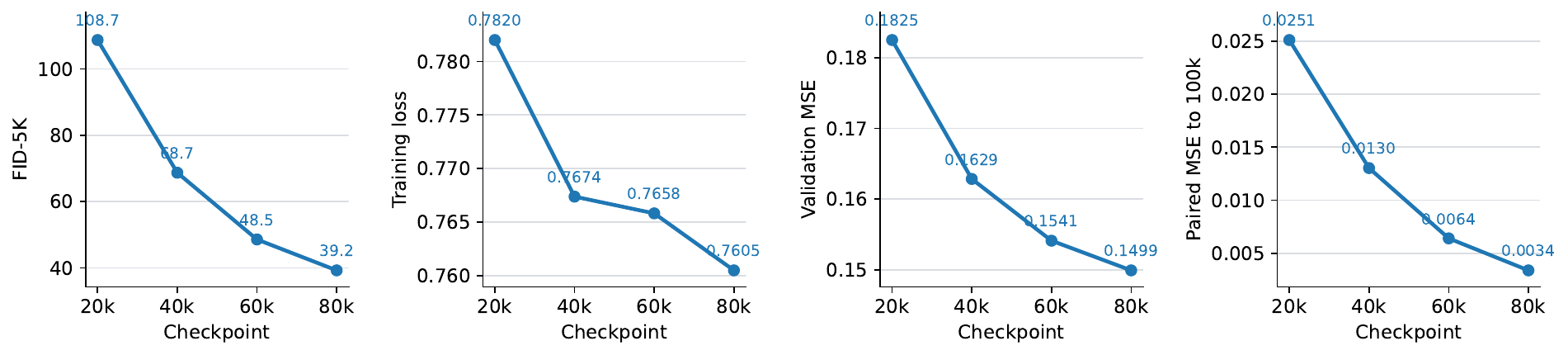}
    \caption{Checkpointwise baseline trends for FID, training loss,
    validation MSE, and paired MSE to the final-checkpoint output. In regression learning, 
    true risk includes empirical risk and a generalization gap.
    Early checkpoints are assumed to have higher true risk, as they tend to have higher
    training error (i.e., empirical risk) and a population risk (i.e., approximated by
    MSE with respect to 
    the validation dataset). We observe that the checkpoints with potentially high true 
    risk are associated with worse FID, supporting our speculated FID-risk relationship.
    Alternatively, if we treat the final 100k-step checkpoint output as the ground-truth
    ideal generation, the same speculated relationship is still supported.}
    \label{fig:baseline-intermediate-complexity}
\end{figure*}

We first define the complexity measure studied in this work: Rademacher complexity.
\begin{definition}[Rademacher complexity]
Let $S=(x_1,\ldots,x_n)$ be a sample and let $\mathcal{H}$ be a class of
real-valued functions. For independent Rademacher variables
$\sigma_1,\ldots,\sigma_n$, the empirical Rademacher complexity of
$\mathcal{H}$ on $S$ is
\begin{equation}
\widehat{\mathfrak{R}}_S(\mathcal{H})
=
\mathbb{E}_{\boldsymbol{\sigma}}
\left[
\sup_{h\in\mathcal{H}}
\frac{1}{n}\sum_{i=1}^{n}\sigma_i h(x_i)
\right],
\;
\sigma_i\sim\operatorname{Unif}\{-1,+1\}.
\label{eq:rademacher-complexity-background}
\end{equation}
The expected Rademacher complexity is
$\mathfrak{R}_n(\mathcal{H})=
\mathbb{E}_{S}[\widehat{\mathfrak{R}}_S(\mathcal{H})]$.
\end{definition}

Intuitively, Rademacher complexity measures the complexity of a function class output by
calculating its worst-case correlation with random noise (i.e., Rademacher variables).
For example, a predictor class capable of generating arbitrary black-and-white noise 
patterns has higher Rademacher complexity than a class restricted to simple 
patterns, such as a black square on a white background, 
due to the former picture's higher worst-case per-pixel correlation with random noise. 
Rademacher complexity can be applied to measure both predictor-output
complexity and loss-function output complexity.

Meanwhile, the down-up sampling interpolation in
Eq.~\eqref{eq:flow-roughening-map} attenuates high-frequency image
components. Theorem~\ref{prop:independent-pixel-benchmark} formalizes the
resulting reduction in coordinate-matching complexity for a general bounded
vector-valued predictor class.
Importantly, Rademacher complexity can directly form the upper bound
of expected regressor true risk; higher Rademacher complexity can
worsen the expected regressor true risk bound, under the assumption that 
the virtual ideal generation data distribution has the same expected mean 
as the training data distribution. 
By true risk, we refer to the actual regression error between 
the generated images and a virtual ideal generation target associated 
with each generated image. Intuitively, if the difference (e.g., regression error)
between the actual generation and the ideal target is small, 
the generation is ideal. Therefore, we associate the generalization true risk 
with FID \citep{heusel2017gans}. Beyond this intuition, we also present a
mechanism study that demonstrates the positive 
correlation between the true risk and FID in
Figure~\ref{fig:baseline-intermediate-complexity}.

Because the roughening operator produces intermediate-step
outputs of lower Rademacher complexity and thus potentially lower
risk (i.e., error), we speculate that some intermediate high-frequency information 
in standard flow matching may be redundant and even harmful, exacerbating 
high-frequency artifacts during sampling. In contrast, the roughening operator can 
improve the final output quality, as reducing the down-up sampling roughening extent 
progressively still recovers a fine image at the end, while the intermediate 
steps involve lower complexity and associated risks. The intuition 
is similar to spending less effort during the intermediate stages of an oil painting.

%% file: sections/4_method.tex
\section{Stroke-Size Control as an Intervention}
\label{sec:method}

We consider conditional image generation with images
$x_0\in\mathbb{R}^{C\times H\times W}$ and class labels
$c\in\{1,\ldots,K_{\mathrm{cls}}\}$. We use a discrete flow-matching path
with $N$ steps and integrate from noise to data on the corresponding
reverse-time grid. For equal-shaped per-example tensors $u$ and $v$, let
$\operatorname{MSE}(u,v)=d_u^{-1}\|u-v\|_2^2$, where $d_u$ is the number
of scalar coordinates. This matches the coordinate-mean reduction used by
the implemented training objectives. Section~\ref{sec:theory} explicitly
uses an unnormalized coordinate-sum surrogate instead.

\subsection{MultiStroke Flow Matching}
\label{sec:multistroke-flow}
In short, MultiStroke coarsens the supervision target of each 
flow-matching time step, where the high-noise time steps are 
associated with rougher targets. We expect the roughening 
to simplify the target and regularize the training. 
Sampling is updated with a roughness-transition residual 
corrector to match the change in training target.  

Let $N\geq2$, set $h=1/N$, let $I$ denote the identity operator, and let
$P_b$ denote the operator that averages each non-overlapping $b\times b$
spatial block independently in every channel and then replicates each
block mean. At time $t\in[h,1]$ along the flow-matching path, the
pixel-space roughening map is
\begin{equation}
    R_t=(1-t+h)I+(t-h)P_b.
    \label{eq:flow-roughening-map}
\end{equation}
Thus $R_h=I$, while $R_1$ retains a fine-scale coefficient $h$.
Because $P_b$ is a projection, the same map can be written as
\begin{equation}
R_t=P_b+(1-t+h)(I-P_b).
\label{eq:flow-roughening-decomposition}
\end{equation}
Hence $R_t$ is invertible for every $t\in[h,1]$. 

Section~\ref{sec:oracle-consistency} formalizes this schedule as a
constant-velocity flow and establishes its non-clairvoyant correction.

\paragraph{Velocity-field training.}
We sample $u\sim\operatorname{Unif}[0,1]$, set $t=\max(u,h)$, and draw an
independent Gaussian latent $\epsilon\sim\mathcal{N}(0,I)$. Using a posterior
sample from the frozen VAE encoder, the roughened data endpoint and the
interpolated latent state are
\begin{equation}
\begin{aligned}
z_t &= 0.18215\,\operatorname{Enc}_{\mathrm{sample}}(R_t(x_0)),\\
x_t &= (1-t)z_t+t\epsilon,\qquad
u_t = \epsilon-z_t.
\end{aligned}
\label{eq:multistroke-training-path}
\end{equation}
Here $\operatorname{Enc}_{\mathrm{sample}}$ denotes posterior sampling from
the frozen VAE encoder, $\operatorname{Enc}_\mu$ its posterior mean, and
$\operatorname{Dec}$ the frozen VAE decoder.
The clamping operation places probability mass $h$ at $t=h$ rather than
sampling exactly uniformly on $[h,1]$. With $N=100$, this affects $1\%$ of
draws and constitutes a minor implementation discrepancy relative to the
baseline's continuous uniform time sampling; its practical effect is expected
to be negligible.
The class-conditional velocity model is optimized with
\begin{equation}
\mathcal{L}_v(\theta)
=\mathbb{E}_{x_0,c,t,\epsilon}
\left[\operatorname{MSE}\left(v_\theta(x_t,t,c),u_t\right)\right].
\label{eq:multistroke-velocity-loss}
\end{equation}
Thus, at each sampled time, the model regresses the ordinary linear-path
velocity for the corresponding roughened endpoint. Because $z_t$ itself
changes with $t$, this target omits the adjacent rough-to-fine endpoint
change; the corrector below supplies that change during sampling.

\paragraph{Adjacent-transition corrector.}
The VAE nonlinearity prevents the exact pixel-space transition from being
applied analytically in latent space. We therefore train a compact residual
corrector separately. Sample
$t\sim\operatorname{Unif}\{2h,3h,\ldots,1\}$ and set $t'=t-h$. To match the
deterministic correction target, define posterior-mean latents and their
adjacent residual by
\begin{equation}
\begin{aligned}
z_t^\mu &= 0.18215\,\operatorname{Enc}_\mu(R_t(x_0)),\\
z_{t'}^\mu &= 0.18215\,\operatorname{Enc}_\mu(R_{t'}(x_0)),\\
\Delta_t^\mu &= z_{t'}^\mu-z_t^\mu.
\end{aligned}
\label{eq:multistroke-corrector-target}
\end{equation}
The corrector receives only the coarser latent and both levels; it never
observes $x_0$ or $\epsilon$. It minimizes
\begin{equation}
\mathcal{L}_c(\phi)
=\mathbb{E}_{x_0,t}
\left[\operatorname{MSE}\left(
c_\phi(z_t^\mu,t,t'),\Delta_t^\mu
\right)\right].
\label{eq:multistroke-corrector-loss}
\end{equation}
This objective approximates the VAE-induced transition separately from
velocity-model training and requires only adjacent level pairs.

\paragraph{Corrected reverse-Euler sampling.}
Initialize $x_1\sim\mathcal{N}(0,I)$. At a grid point $t=jh$, the velocity is
$v_t=v_\theta(x_t,t,c)$.
For a nonfinal step, set $t'=t-h$. The current clean-endpoint estimate and
the complete corrected update are
\begin{equation}
\begin{aligned}
\widehat z_t &= x_t-tv_t,\qquad
\widehat\Delta_t = c_\phi(\widehat z_t,t,t'),\\
x_{t'} &= x_t-hv_t+(1-t')\widehat\Delta_t.
\end{aligned}
\label{eq:multistroke-corrected-euler}
\end{equation}
The first two terms are the reverse-Euler update. The last term changes only
the clean-endpoint contribution: if the predicted residual equals
$\Delta_t=z_{t'}-z_t$, its coefficient $1-t'$ replaces the contribution of
$z_t$ by that of $z_{t'}$ while preserving the Gaussian component. This is
the oracle identity analyzed in Section~\ref{sec:oracle-consistency}. At the
final step $t=h$, no correction is needed because $R_h=I$; the generated
latent and image are
\begin{equation}
z_{\mathrm{out}}=x_h-hv_h,
\qquad
\widehat x_0=\operatorname{Dec}(z_{\mathrm{out}}/0.18215).
\label{eq:multistroke-final-step}
\end{equation}
Appendix~\ref{app:flow-impl} provides the procedure summary in
Algorithm~\ref{alg:multistroke-flow-corrector} and the complete specification
in Algorithm~\ref{app:algorithm-multistroke-flow}.

%% file: sections/4_theory.tex
\section{Theoretical Analysis}

\label{sec:theory}

\subsection{Consistency of Corrected Sampling}
We analyse the consistency of corrected sampling.
\label{sec:oracle-consistency}

The learned corrector $c_\phi$ is trained on posterior-mean VAE latents of
training images at adjacent roughness levels to predict their residual transition. The
corresponding oracle residual yields an exact transition along the prescribed
path.

Following Section~\ref{sec:multistroke-flow}, let $x_0$ be a clean image,
$c$ its class label, $h=1/N$, $r_t=R_t(x_0)$, and
$\epsilon\sim\mathcal N(0,I)$ independent noise. Set $t'=t-h$ and define
$X_t=(1-t)r_t+t\epsilon$, $U_t=\epsilon-r_t$, and the population MSE
regressor $v^*(x,t,c)=\mathbb E[U_t\mid X_t=x,c]$.

\begin{proposition}[Constant velocity and non-clairvoyant correction]
\label{prop:constant-roughness-velocity}
The roughness path has constant velocity 
$\dot r_t=-(I-P_b)x_0$. For $t\in[2h,1]$, its adjacent reverse displacement
is recoverable from the current endpoint, and its conditional mean satisfies
\begin{equation}
\begin{aligned}
r_{t'}-r_t
&=\frac{h}{1-t+h}(I-P_b)r_t,\\
\mathbb E[r_{t'}-r_t\mid X_t=x,c]
&=\frac{h}{1-t+h}(I-P_b)
\bigl(x-tv^*(x,t,c)\bigr).
\end{aligned}
\label{eq:ms-nonclairvoyant-transition}
\end{equation}
Hence the exact population correction depends only on the observable flow
state, time, and condition, not on the hidden pair $(x_0,\epsilon)$.
\end{proposition}

\begin{proof}
Differentiate Eq.~\eqref{eq:flow-roughening-decomposition}. Then use
$(I-P_b)r_t=(1-t+h)(I-P_b)x_0$ and
$x-tv^*(x,t,c)=\mathbb E[r_t\mid X_t=x,c]$.
\end{proof}

The frozen nonlinear VAE generally maps this \textbf{straight} pixel-space path to a
nonconstant latent trajectory. Fix a coupled VAE latent realization across
time, write $\operatorname{Enc}$ for that realized encoder map, and use the
same $\epsilon$. With the encoder scaling included, the latent path is
\begin{equation}
\begin{aligned}
R_t &= (1-t+h)I+(t-h)P_b,\\
z_t &= 0.18215\,\operatorname{Enc}(R_t(x_0)),\\
x_t &= (1-t)z_t+t\epsilon,\qquad t\in[h,1].
\end{aligned}
\label{eq:ms-oracle-path}
\end{equation}
For a nonfinal grid step $t=jh$ ($j\ge2$),
$u_t=\epsilon-z_t$, and $\Delta_t=z_{t'}-z_t$. The time-dependent clean
endpoint gives
\begin{equation}
\begin{aligned}
\dot x_t
&=\frac{d}{dt}\bigl((1-t)z_t+t\epsilon\bigr)\\
&=-z_t+(1-t)\dot z_t+\epsilon
=u_t+(1-t)\dot z_t.
\end{aligned}
\label{eq:ms-oracle-velocity}
\end{equation}
The standard target $u_t$ omits $(1-t)\dot z_t$; the separately trained
corrector supplies the resulting adjacent-level transition during sampling.
The oracle-corrected update satisfies
\begin{equation}
x_t-hu_t+(1-t')\Delta_t
=(1-t')z_{t'}+t'\epsilon=x_{t'}.
\label{eq:ms-oracle-step}
\end{equation}
Moreover, the oracle endpoint estimate and final step obey
\begin{equation}
\begin{aligned}
x_t-tu_t
&=(1-t)z_t+t\epsilon-t(\epsilon-z_t)=z_t,\\
x_h-hu_h
&=(1-h)z_h+h\epsilon-h(\epsilon-z_h)=z_h,
\end{aligned}
\label{eq:ms-oracle-endpoints}
\end{equation}
where $R_h=I$. Oracle predictions therefore recover every grid point and the
unroughened endpoint exactly.

\subsection{Generalization Bounds Analysis}
\label{sec:ms-complexity}
Each step's population risk is bounded by its empirical loss and Rademacher
complexity. We then quantify how roughening reduces coordinate-matching
complexity.

\begin{assumption}[Shared ghost target]
\label{ass:shared-ghost-target}
At each sampling level, the baseline and MultiStroke risks are evaluated
against the same ideal ground-truth target. 
\end{assumption}

\paragraph{Setup.}
Let $S=\{(x_i,y_i)\}_{i=1}^{n}$ be an i.i.d.\ sample drawn from a
distribution $\mathcal{D}$, where $y_i$ is the shared ghost target from
Assumption~\ref{ass:shared-ghost-target}. Each target
$y_i\in\mathbb{R}^{d}$ contains $d=CHW$ scalar output coordinates, including
channels, and a predictor $h\in\mathcal{H}$ produces
$h(x)=(h_1(x),\ldots,h_d(x))$. We consider the coordinate-sum MSE loss
\begin{equation}
\ell(h(x),y)
=
\sum_{p=1}^{d}
\bigl(h_p(x)-y_p\bigr)^2.
\label{eq:pixel_mse}
\end{equation}
The empirical risk is
\begin{equation}
\widehat{R}_S(h)=\frac{1}{n}\sum_{i=1}^{n}\ell(h(x_i),y_i).
\label{eq:risk_def}
\end{equation}
The population risk is $R(h)=\mathbb{E}_{(X,Y)\sim\mathcal{D}}
[\ell(h(X),Y)]$.
Assume $|h_p(x)|\leq B$ and $|y_p|\leq Y$ for all $p$.

For independent Rademacher variables
$\sigma_{ip}\sim\mathrm{Unif}\{-1,+1\}$, independent of the sample $S$,
let $\boldsymbol\sigma_i=(\sigma_{i1},\ldots,\sigma_{id})^\top$ and define
\begin{equation}
\begin{aligned}
\widehat{\mathfrak R}^{\mathrm{vec}}_S(\mathcal H)
&=\frac1n\mathbb E_{\sigma}\left[
\sup_{h\in\mathcal H}\sum_{i=1}^n
\boldsymbol\sigma_i^\top h(x_i)\right],\\
\mathfrak R^{\mathrm{vec}}_n(\mathcal H)
&=\mathbb E_S\widehat{\mathfrak R}^{\mathrm{vec}}_S(\mathcal H).
\end{aligned}
\label{eq:vector_rad_def}
\end{equation}
For a scalar function class $\mathcal G$, let
$\widehat{\mathfrak R}_S(\mathcal G)$ denote the analogous empirical
complexity and $\mathfrak R_n(\mathcal G)$ its expectation over $S$.

\begin{lemma}[Generalization under coordinate-sum MSE]
\label{lem:mse_rad}
Let $\mathcal{L}=\{(x,y)\mapsto\ell(h(x),y):h\in\mathcal{H}\}$ be the loss class induced by $\mathcal{H}$. Then any data-dependent predictor $\widehat{h}\in\mathcal{H}$ satisfies
\begin{equation}
\mathbb E_S[R(\widehat h)]
\leq
\mathbb E_S[\widehat R_S(\widehat h)]
+
4\sqrt{2d}(B+Y)
\mathfrak R_n^{\mathrm{vec}}(\mathcal H).
\label{eq:mse_gen_bound}
\end{equation}
\end{lemma}

The proof is given in Appendix~\ref{app:regression-proof}.
Thus empirical loss and complexity enter the same step-wise bound.

\paragraph{Reduced coordinate matching.}

\begin{theorem}[Coordinate matching under block roughening]
\label{prop:independent-pixel-benchmark}
Let $A$ average each disjoint block of $m\geq2$ output coordinates and
replicate the block mean at every coordinate, where $m$ divides $d$. Define
$T_\alpha=(1-\alpha)I+\alpha A$ for $\alpha\in[0,1]$. Let the inputs
$x_1,\ldots,x_n$ in $S$ be
pairwise distinct, and consider the fully coordinate-flexible class
$\mathcal H_{\square}=\{h:\mathcal X\to[-B,B]^d\}$, where $\mathcal X$ is
the input domain.
For independent Rademacher variables $\sigma_1,\ldots,\sigma_m$,
\begin{equation}
\begin{aligned}
\frac{1}{d}\widehat{\mathfrak R}^{\mathrm{vec}}_S
(T_\alpha\mathcal H_{\square})
&= B\,\mathbb E_\sigma\left|
(1-\alpha)\sigma_1+
\frac{\alpha}{m}\sum_{j=1}^m\sigma_j\right|\\
&\leq B\sqrt{1-(2\alpha-\alpha^2)
\left(1-\frac1m\right)}.
\end{aligned}
\label{eq:independent_pixel_bound}
\end{equation}
At complete block averaging, the exact normalized complexity ratio is
\begin{equation}
\frac{\widehat{\mathfrak R}^{\mathrm{vec}}_S
(A\mathcal H_{\square})}
{\widehat{\mathfrak R}^{\mathrm{vec}}_S(\mathcal H_{\square})}
=\frac{1}{m}\mathbb E_\sigma
\left|\sum_{j=1}^m\sigma_j\right|
\leq\frac{1}{\sqrt m}.
\label{eq:independent_pixel_ratio}
\end{equation}
For even $m$, the ratio equals $2^{-m}\binom{m}{m/2}$;
for $2\times2$ and $8\times8$ blocks it is, respectively,
$3/8$ and $2^{-64}\binom{64}{32}\approx0.0993$.
\end{theorem}

The proof is provided in Appendix~\ref{app:block-average-proof}. Averaging
forces each block value to align with a sum of independent signs rather than
with each coordinate separately. Note that the predictor here 
corresponds to the general-architecture predictor with image output during sampling, 
instead of one with just gaussian noise prediction.

%% file: sections/6_exp.tex
\section{Experiments and Findings}
\label{sec:experiments}

\paragraph{Empirical complexity of the surrogate task.}
We construct paired linear flow-matching surrogates from ImageNet validation
images at $256\times256$ resolution. The flow time $t$ and $8\times8$
roughening weight $\alpha$ are sampled independently, making this a mechanism
probe rather than a reproduction of the schedule $\alpha=t-h$. Roughening
changes the clean endpoint used for both the noisy input and velocity target;
the paired conditions otherwise share images, times, noise, output
coordinates, and Rademacher signs. We use eight disjoint subsets of 64 images,
256 sign draws, and 32 output coordinates.

Let $D$ be the input dimension, $\operatorname{vec}(x)\in\mathbb R^D$ its
vectorization, $(1,t)\in\mathbb R^2$ a bias-augmented time feature, and
$\otimes$ the Kronecker product. Define
\begin{equation}
\phi(x,t)=\frac{\operatorname{vec}(x)}{\sqrt D}
\otimes\frac{(1,t)}{\sqrt{1+t^2}}.
\label{eq:surrogate-feature-map}
\end{equation}
For a subset $S=\{(x_i,t_i,y_i)\}_{i=1}^n$, selected coordinates $\mathcal Q$,
coordinate weights $w_q\in\mathbb R^{2D}$, and independent signs
$\sigma_i\sim\operatorname{Unif}\{-1,+1\}$, we estimate the squared-loss
complexity by
\begin{equation}
\widehat{\mathfrak C}_S
=\frac{1}{|\mathcal Q|n}\mathbb E_\sigma
\sum_{q\in\mathcal Q}\sup_{\|w_q\|_2\leq1}
\sum_{i=1}^n\sigma_i
\left[\bigl(w_q^\top\phi(x_i,t_i)-y_{iq}\bigr)^2-y_{iq}^2\right].
\label{eq:target-aware-complexity}
\end{equation}
Subtracting $y_{iq}^2$ leaves the exact Rademacher complexity unchanged but
removes hypothesis-independent Monte Carlo noise. The resulting loss-class
estimate is target-aware. It is an empirical mechanism probe rather than an
assumption of the general theory, which permits an arbitrary bounded
vector-valued predictor.

\input{tables/rademacher_surrogate}

\paragraph{Surrogate-complexity results.}
Table~\ref{tab:rademacher-surrogate} shows that roughening reduces the input
norm by $1.33\%\pm0.14\%$, the target norm by $1.04\%\pm0.10\%$, and
empirical complexity by $1.99\%\pm0.17\%$. All three quantities decrease on
every subset. The reduction in empirical complexity is modest but consistent
and exceeds the reductions in either norm, providing controlled evidence for
the proposed lower-complexity mechanism. This linear-surrogate estimate does
not characterize the Rademacher complexity of the latent transformer and is
not required by the theoretical results.

\paragraph{Main experiment setup.}
We compare the vanilla flow-matching model \citep{lipman2023flow} with
MultiStroke on ImageNet-1K \citep{deng2009imagenet} at $256\times256$
resolution. Both models use the same latent transformer, data pipeline,
optimization settings, training duration, and global batch size. We evaluate
MultiStroke with $2\times2$ and $8\times8$ pixel-space roughening; both
variants use an adjacent-transition corrector during sampling. Generation
uses 100 Euler steps for all methods.
We evaluate the $2\times2$ and $8\times8$ settings because the roughening
factor should divide the image dimensions to form complete averaging blocks,
and overly aggressive roughening may drastically reduce the learning signal.

We report FID \citep{heusel2017gans},
class-conditioned CLIP similarity \citep{radford2021learning}, and precision,
recall, density, and coverage \citep{naeem2020reliable}. The two methods use
matched conditioning and initial-noise assignments. FID uses each complete
generated set, whereas CLIP and the distributional metrics use fixed,
identifier-matched subsets. FID and the distributional metrics use a shared
ImageNet validation reference. Appendix~\ref{app:flow-impl} provides complete
implementation and evaluation details.

\input{tables/flow_completed}

\paragraph{Main results.}
The $2\times2$ and $8\times8$ variants reduce FID from $20.17$ to $18.31$
and $15.64$, corresponding to relative improvements of $9.2\%$ and $22.5\%$,
respectively. Class-conditioned CLIP similarity remains comparable across all
methods. The $2\times2$ variant improves precision, recall, density, and
coverage over the baseline. The $8\times8$ variant further improves recall
and coverage, while incurring a modest reduction in precision. These results
suggest that both corrected MultiStroke variants improve distributional quality,
with the stronger roughening producing the largest FID reduction without a
material loss of class-conditioned alignment.

\begin{figure*}[ht]
    \centering
    \includegraphics[width=0.8\linewidth]{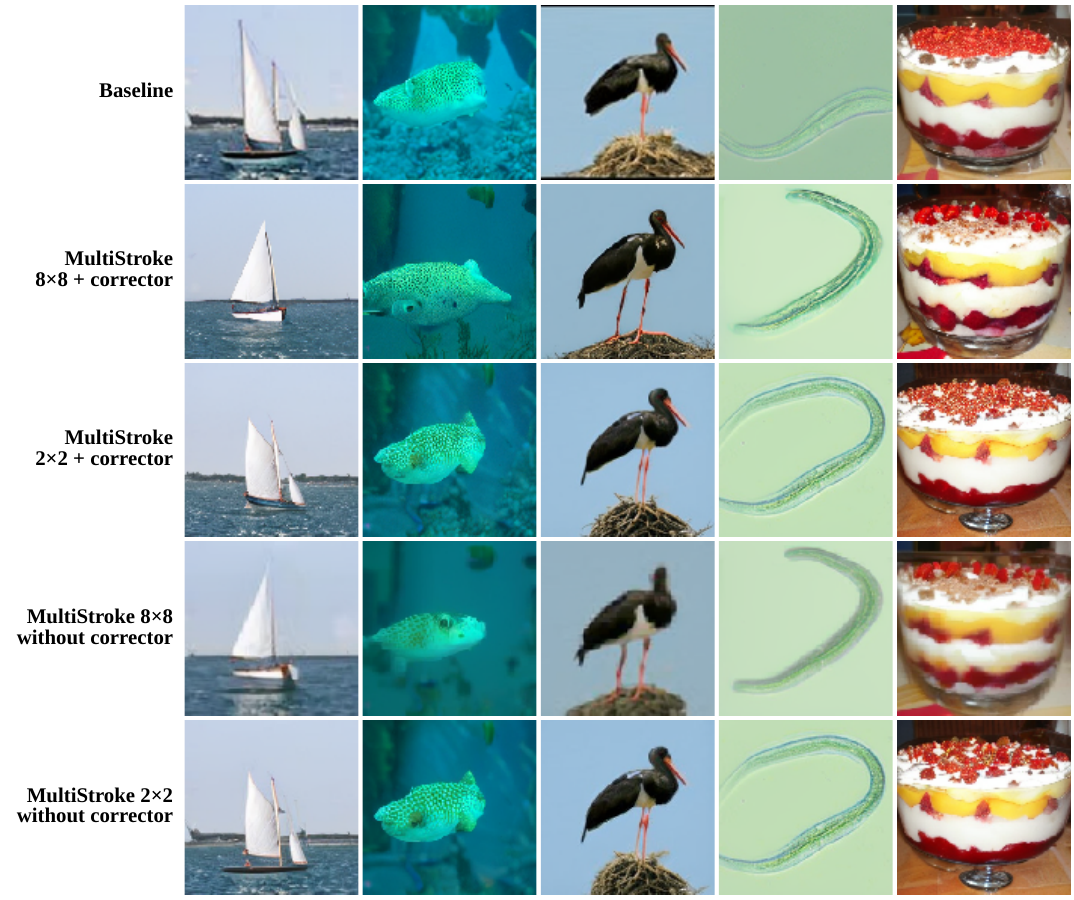}
    \caption{Qualitative comparison using matched class conditions and
    initial noise. Columns correspond to the same five sample identities
    across methods. Rows compare the baseline and
    $2\times2$ and $8\times8$ MultiStroke with and without the
    adjacent-transition corrector. Best viewed in color with zoom.}
    \label{fig:corrector-qualitative}
\end{figure*}

\paragraph{Ablation studies and hyperparameter tuning.}
Changing the roughening stroke size from $8\times8$ to $2\times2$ increases
FID from $15.64$ to $18.31$. This result identifies roughening stroke size as
an important hyperparameter, with larger strokes providing stronger
distributional improvement in this setting. We further ablate the transition
corrector for the $2\times2$ model. Without the corrector, MultiStroke attains
an FID of $19.35$, improving upon the baseline FID of $20.17$ but trailing the
corrected FID of $18.31$. The corrector therefore remains effective. At the
same time, the improvement without correction persists despite the
granularity mismatch between adjacent roughness levels during sampling,
providing further evidence that varying stroke size itself improves the
learned generative path. This $2\times2$ quantitative ablation already
demonstrates the degradation caused by removing the corrector. We do not
report distributional metrics for the $8\times8$ model without correction
because its generations exhibit an obvious checkerboard failure mode; the
matched qualitative grid in Figure~\ref{fig:corrector-qualitative} documents
this failure directly.

We additionally quantify the approximation introduced by the transition
corrector using the separately trained 10k-update correctors for $2\times2$
and $8\times8$ pixel roughening. Figure~\ref{fig:corrector-qualitative}
provides a matched qualitative comparison. Without the corrector,
$8\times8$ MultiStroke exhibits pronounced checkerboard artifacts, which the
corrector visibly mitigates.
We also evaluate the first
100 ImageNet validation images at
the adjacent transitions $100\rightarrow99$, $75\rightarrow74$,
$50\rightarrow49$, and $25\rightarrow24$. For each image and transition,
we compare the true finer-level posterior-mean VAE latent with the
coarser-level latent before correction and with the predicted latent after
correction, then average the per-image latent MSE. For $2\times2$
roughening, the before/after MSEs are
$4.548\times10^{-4}/4.511\times10^{-4}$,
$3.509\times10^{-4}/3.427\times10^{-4}$,
$2.739\times10^{-4}/2.680\times10^{-4}$, and
$2.523\times10^{-4}/2.505\times10^{-4}$, respectively. For $8\times8$
roughening, they are $7.967\times10^{-4}/7.816\times10^{-4}$,
$1.626\times10^{-3}/1.537\times10^{-3}$,
$1.283\times10^{-3}/1.234\times10^{-3}$, and
$6.234\times10^{-4}/6.020\times10^{-4}$.
The corrector lowers mean latent MSE at every evaluated transition, with
relative reductions between $0.7\%$ and $5.5\%$. These modest reductions
suggest that the corrector improves average transition accuracy but only
partially compensates for the nonlinear VAE transition, leaving measurable
transition prediction error; a more accurate
corrector could further improve sampling.

\paragraph{Additional studies.}
For latent-only $2\times2$ MultiStroke, let $A_2$ average each non-overlapping
$2\times2$ latent block and replicate its mean, let $h=1/N$, and define
$a_t=\max(t,h)-h$ and $L_t=(1-a_t)I+a_tA_2$. Given a clean VAE latent $z$
and Gaussian endpoint $\epsilon$, training uses
\begin{equation}
x_t=(1-t)L_tz+t\epsilon,
\qquad
u_t=\epsilon-L_tz.
\label{eq:latent-only-multistroke}
\end{equation}
During sampling, a velocity prediction $\widehat u_t$ gives the rough clean
endpoint $\widehat z_t^{\mathrm r}=x_t-t\widehat u_t$ and Gaussian endpoint
$\widehat\epsilon_t=x_t+(1-t)\widehat u_t$. Since $A_2$ is idempotent, the
inverse roughening operator is
\begin{equation}
L_t^{-1}y=A_2y+\frac{(I-A_2)y}{1-a_t},
\;
x_{t'}=(1-t')L_{t'}L_t^{-1}\widehat z_t^{\mathrm r}
+t'\widehat\epsilon_t.
\label{eq:latent-only-inverse-transition}
\end{equation}
This variant is inferior to pixel-space MultiStroke. A likely cause is that
VAE features are nonlinear: directly roughening latent features disrupts the
intended constant-velocity field transition and may introduce undesirable
distortions.

We also study corrector stochasticity. For adjacent roughness levels, write
the two VAE posteriors as
$q_t=\mathcal N(\mu_t,\operatorname{diag}(\sigma_t^2))$ and
$q_{t'}=\mathcal N(\mu_{t'},\operatorname{diag}(\sigma_{t'}^2))$. We draw one
$\xi\sim\mathcal N(0,I)$ and use the shared reparameterization
\begin{equation}
z_t^\xi=\mu_t+\sigma_t\odot\xi,
\qquad
z_{t'}^\xi=\mu_{t'}+\sigma_{t'}\odot\xi,
\qquad
\Delta_t^\xi=z_{t'}^\xi-z_t^\xi.
\label{eq:shared-posterior-corrector}
\end{equation}
The corrector minimizes
$\mathbb E\|c_\phi(z_t^\xi,t,t')-\Delta_t^\xi\|_2^2$. Reusing the same
posterior noise at both levels prevents independent sampling noise from
entering the adjacent residual. This formulation closely matches the
deterministic VAE corrector, which replaces both reparameterized samples by
their posterior means.

Finally, the multiscale-loss baseline retains the ordinary latent path and
velocity target but weights each example by the MultiStroke fine-detail
schedule,
\begin{equation}
\mathcal L_{\mathrm{multi}}
=\mathbb E\left[(1-\max(t,h)+h)
\|v_\theta(x_t,t,c)-(\epsilon-z)\|_2^2\right].
\label{eq:multiscale-loss-baseline}
\end{equation}
This control improves FID-5K from $29.73$ to $28.72$, but the gain is smaller
than when the block-averaged components are added. Because its weighting
approximates the most aggressive attenuation of fine-detail supervision, the
result suggests that MultiStroke is unlikely to degrade baseline performance
even at the greatest roughening extent. Relative to this control, adding the
roughened components improves FID-5K by $25.4\%$ and $17.4\%$ for the main
$8\times8$ and $2\times2$ MultiStroke variants, respectively. This result
supports the intuition that a coarse layout helps the model predict fine
components more easily.

\paragraph{Corrector overhead and comparison with other works.}
The generator and frozen VAE together contain $535.70$M parameters, while the
transition corrector contains only $77{,}476$ parameters. The corrector thus
accounts for only approximately $0.014\%$ of the combined parameter budget.
It is supervised only on the latent counterpart of the deterministic,
non-clairvoyant transition in
Proposition~\ref{prop:constant-roughness-velocity}; it receives no class,
clean-image, noise, or FID information. Therefore, while 
we cannot fully rule out the single confound on corrector, it is extremely unlikely
that the MultiStroke performance improvement is attributable to the
minimal component or its supervision.

Table~\ref{tab:matched-training-exposure} compares MultiStroke with prior
methods under the same training exposure. We restrict this comparison to
systems comprising a generator and VAE, with generative backbones trained
from random initialization and no auxiliary large pretrained encoder or
representation model. Under this protocol, MultiStroke attains an ADM
FID~\citep{dhariwal2021diffusion}
of $13.71$ and $13.72$ with the
stochastic and deterministic correctors, respectively, both of which are
superior to those of most prior works.

\paragraph{Computational cost.}
The ImageNet experiments are conducted on eight
NVIDIA H200 GPUs. Under matched configurations, the baseline and MultiStroke velocity
models each completed 100k updates in approximately 9.9 hours. Training the
lightweight adjacent-transition corrector requires only frozen VAE encoding
(i.e., excluding the generator) and takes approximately 20 minutes for 10k updates.
Under the matched evaluation workload, image generation required approximately
1.5 hours for both the baseline and MultiStroke with the corrector. These measurements indicate comparable
velocity-model training and sampling time, with modest additional
corrector-training cost. The matched computation budget is
expected due to the lightweight nature of down-up sampling.

\input{tables/matched_training_exposure}

\paragraph{Additional experiments on CIFAR-DDPM.}

We additionally test whether MultiStroke transfers from flow matching to a
class-conditional pixel-space DDPM \citep{ho2020ddpm} on CIFAR-10
\citep{krizhevsky2009learning}. At each
diffusion level, we linearly mix both the ordinary noisy image and its noise
prediction target with their non-overlapping $2\times2$ block averages.
During sampling, each stochastic DDPM update is formed at the current roughness
level, mapped back through the inverse roughening operator, and then mapped
to the next finer selected level.
This construction requires no corrector.

Algorithm~\ref{alg:multistroke-cifar-ddpm} in
Appendix~\ref{app:cifar-ddpm} summarizes training and sampling,
alongside the exact
operators, posterior updates, clipping order, and
implementation settings. We train the baseline
and MultiStroke models for 100k updates with matched architectures,
optimization settings, and global batch sizes, and evaluate both with the
same fixed-small ancestral sampler. Table~\ref{tab:cifar-ddpm} shows that
MultiStroke reduces FID from $29.04$ to $23.16$, an improvement of $20.2\%$.
Appendix Figure~\ref{fig:cifar-ddpm-qualitative} provides the corresponding
qualitative comparison under matched class and initial-noise assignments.
Together, these results provide evidence that roughening the input and
regression target may improve a discrete diffusion model in addition to the
flow-matching models.

\input{tables/cifar_ddpm_results}

%% file: tables/rademacher_surrogate.tex
\begin{table}[t]
\centering
\caption{Target-aware linear-surrogate comparison under sampled $8\times8$
roughening. Entries are means with 95\% confidence intervals across eight
disjoint subsets. Ratios are computed within each subset before aggregation;
values below one favor the roughened task.}
\label{tab:rademacher-surrogate}
\small
\resizebox{\columnwidth}{!}{%
\begin{tabular}{lccc}
\toprule
Metric & Baseline & Roughened & Ratio \\
\midrule
Input norm & $0.6544\pm0.0105$ & $0.6456\pm0.0105$ & $0.9867\pm0.0014$ \\
Target norm & $1.1513\pm0.0045$ & $1.1393\pm0.0043$ & $0.9896\pm0.0010$ \\
Empirical complexity & $0.1955\pm0.0043$ & $0.1916\pm0.0043$ & $0.9801\pm0.0017$ \\
\bottomrule
\end{tabular}
}
\end{table}

%% file: tables/flow_completed.tex
\begin{table}[t]
\centering
\caption{Matched ImageNet-1K flow-matching comparison. MultiStroke uses
pixel-space roughening and a transition corrector trained after the velocity
model. All velocity models use 100k training updates; each corrector uses 10k
additional updates. Parentheses report the relative FID improvement over the
baseline.}
\label{tab:flow-completed}
\small
\resizebox{\columnwidth}{!}{%
\begin{tabular}{lccc}
\toprule
Metric & Baseline & MS $2\times2$ + corr. & MS $8\times8$ + corr. \\
\midrule
FID $\downarrow$ & 20.17 & 18.31 ($9.2\%$) & 15.64 ($22.5\%$) \\
CLIP $\uparrow$ & 26.62 & 26.52 & 26.35 \\
Precision $\uparrow$ & 0.972 & 0.973 & 0.962 \\
Recall $\uparrow$ & 0.584 & 0.592 & 0.618 \\
Density $\uparrow$ & 1.504 & 1.568 & 1.550 \\
Coverage $\uparrow$ & 0.958 & 0.961 & 0.966 \\
\bottomrule
\end{tabular}
}
\end{table}

\begin{table}[t]
\centering
\caption{Matched ImageNet-1K ablation comparison using FID-5K and PRDC. The
$8\times8$ posterior corrector uses shared-noise reparameterization for
adjacent VAE posteriors.}
\label{tab:flow-ablations}
\small
\resizebox{\columnwidth}{!}{%
\begin{tabular}{lccccc}
\toprule
Method & FID-5K $\downarrow$ & Precision $\uparrow$ & Recall $\uparrow$
& Density $\uparrow$ & Coverage $\uparrow$ \\
\midrule
Matched baseline & 29.73 & 0.9812 & 0.5898 & 1.5953 & 0.9700 \\
Latent-only MS $2\times2$ & 29.41 & \textbf{0.9824} & 0.5726
& \textbf{1.6646} & 0.9728 \\
MS $8\times8$ + posterior corrector & \textbf{26.29} & 0.9676
& \textbf{0.5986} & 1.6563 & \textbf{0.9758} \\
Multiscale-loss baseline & 28.72 & 0.9776 & 0.5842 & 1.6554
& 0.9750 \\
\bottomrule
\end{tabular}
}
\end{table}

%% file: tables/matched_training_exposure.tex
\begin{table}[H]
\centering
\caption{ImageNet-1K $256\times256$ results at a matched exposure of 51.2
million examples. All generative backbones start from random initialization;
no method uses a pretrained component beyond its VAE. Parameters are
generator-only/with-VAE, with the MultiStroke corrector included in the
generator count. FID uses the ADM evaluation protocol.}
\label{tab:matched-training-exposure}
\small
\resizebox{\columnwidth}{!}{%
\begin{tabular}{lccc}
\toprule
Method & Sampling & Parameters & FID $\downarrow$ \\
\midrule
EDT-S* \citep{chen2024edt}
& 250 DDIM & \shortstack{38.30M/121.95M\\(exclude/include VAE)}
& \shortstack{55.8/47.0\\(without/with AMM)} \\
EDT-B* \citep{chen2024edt}
& 250 DDIM & 152.00M/235.65M
& \shortstack{37.3/29.0\\(without/with AMM)} \\
SiT-L/2 \citep{wang2026vaerepa}
& 250 Euler-Maruyama & 458.00M/541.65M & 28.20 \\
VAE-REPA (SiT-L/2) \citep{wang2026vaerepa}
& 250 Euler-Maruyama & 458.00M/541.65M & 22.50 \\
EDT-XL* \citep{chen2024edt}
& 250 DDIM & 698.40M/782.05M
& \shortstack{15.6/12.6\\(without/with AMM)} \\
\midrule
MultiStroke $8\times8$
& 100 Euler & 452.13M/535.78M
& \shortstack{13.71/13.72\\(stochastic/deterministic corrector)} \\
\bottomrule
\end{tabular}
}
\end{table}

%% file: tables/cifar_ddpm_results.tex
\begin{table}[t]
\centering
\caption{Matched CIFAR-10 pixel-space DDPM comparison. Both models use the
same architecture, optimization budget, global batch size, and fixed-small
ancestral sampling protocol. Parentheses report the relative FID improvement
over the baseline.}
\label{tab:cifar-ddpm}
\small
\begin{tabular}{lc}
\toprule
Method & FID $\downarrow$ \\
\midrule
DDPM baseline & 29.04 \\
MultiStroke $2\times2$ & 23.16 ($20.2\%$) \\
\bottomrule
\end{tabular}
\end{table}

%% file: sections/7_conclusion.tex
\section{Limitations, Conclusion, and Future Work}
\label{sec:conclusion}
The corrector has training and generalization errors.
Its error can be particularly high at the noisy time steps, 
and increasing the minimum roughness interpolation weight may mitigate 
the problem. Due to the corrector confound, we cannot assert 
a causal mechanism, though the confound is kept minimal. 
Besides, our exploration is
constrained to a basic latent flow-matching setting under a limited computational budget.
Developing a more accurate corrector and establishing broader generality
remain meaningful future work.

In short, in this work, we study stroke-size control as a controlled 
intervention for flow matching. 
MultiStroke coarsens the 
regression target, and then progressively restores fine spatial
detail. Despite the corrector's limited capacity, we achieve
a substantial FID improvement. This work is best viewed
as a mechanism exploration, not a state-of-the-art method paper.

%% file: sections/8_app.tex
\floatname{algorithm}{Algorithm}

\input{sections/2_related_extended}

\section{Flow-Matching Preliminaries}
\label{app:flow-preliminaries}

For the flow-matching model in latent space, let $z_0=\operatorname{Enc}(x_0)$
denote a scaled posterior sample from the frozen VAE for image $x_0$.
Following the
time direction used in the main paper, the linear probability path moves
from the clean latent at time $0$ to a standard Gaussian latent at time
$1$:
\begin{equation}
    x_t=(1-t)z_0+t\epsilon,
    \qquad \epsilon\sim\mathcal{N}(0,I),\quad t\in[0,1].
    \label{eq:flow-forward}
\end{equation}
Its increasing-time velocity is $\epsilon-z_0$. Flow matching
\citep{lipman2023flow} therefore trains the conditional velocity field
with
\begin{equation}
    \mathcal{L}_{\mathrm{FM}}(\theta)=
    \mathbb{E}\left[
        \operatorname{MSE}\!\left(v_\theta(x_t,t,c),\epsilon-z_0\right)
    \right].
    \label{eq:flow-matching-loss}
\end{equation}
For a grid of $N$ steps, let $h=1/N$. Generation starts from
$x\sim\mathcal{N}(0,I)$ at time $s=1$ and integrates this
increasing-time velocity backward with Euler updates,
\begin{equation}
    x\leftarrow x-h\,v_\theta(x,s,c),
    \qquad s=1,1-h,\ldots,h.
    \label{eq:flow-baseline-sampling}
\end{equation}
The final latent is decoded by the VAE. MultiStroke applies its
time-dependent roughening before VAE encoding and uses a learned adjacent
transition corrector during sampling, as described in
Section~\ref{sec:method}.

\clearpage
\section{Flow-Matching Implementation Details}
\label{app:flow-impl}

The flow-matching implementation follows the latent-diffusion
codebase of \citet{gmongaras2023latentdiffusion}.

\paragraph{Dataset and architecture.}
We train class-conditional flow-matching models in latent space on ImageNet-1K
\citep{deng2009imagenet} at $256\times256$ resolution. Both the baseline and
MultiStroke models use the same latent transformer \citep{vaswani2017attention}:
four VAE-latent channels, patch size $2$, width $1088$, $17$ transformer
blocks, $17$ attention heads, and $1000$ class labels. The velocity transformer
has 452.05 million trainable parameters. Including the frozen
83.65-million-parameter VAE, the model contains 535.70 million parameters in
total. The separately trained MultiStroke transition corrector adds 0.077
million trainable parameters.

\paragraph{VAE and latent convention.}
We use the frozen Hugging Face \text{stabilityai/sd-vae-ft-ema}
\text{AutoencoderKL} checkpoint (revision
\text{f04b2c4b98319346dad8c65879f680b1997b204a}) from the latent-diffusion
model \citep{rombach2022ldm}. Images are normalized
to $[-1,1]$ before encoding. The default method draws a VAE posterior
sample and applies the VAE's native shift and scaling factors. For this
checkpoint, the shift factor is absent and
$z=0.18215\,\operatorname{Enc}_{\mathrm{sample}}(x)$. During generation, we
decode $\operatorname{Dec}(z/0.18215)$ and clamp the output to $[-1,1]$.
The VAE is fixed for both baseline and MultiStroke models.

\paragraph{Optimization and sampling.}
Both velocity models are trained from scratch for 100{,}000 updates on eight
GPUs, using 64 images per GPU and a global batch size of 512. We use AdamW
\citep{loshchilov2019decoupled}
with learning rate $10^{-4}$, weight decay $0.01$, and bfloat16
mixed-precision distributed training. Both
methods use the final exponential-moving-average checkpoint. Generation uses
100 Euler steps and identical class and initial-noise
assignments for corresponding image identifiers. FID \citep{heusel2017gans}
uses the StyleGAN
Inception 2015-12-05 feature extractor with 2048-dimensional features and a
shared ImageNet validation reference.

\paragraph{Training-loss comparison.}
MultiStroke achieves lower training loss than the baseline, with greater
roughening producing a larger reduction. Because the velocity objective is
the coordinate-averaged squared velocity prediction error, this result means
that MultiStroke reduces the average squared velocity prediction error on the
training data. In Lemma~\ref{lem:mse_rad}, this reduction lowers the empirical
risk term of the generalization bound when the complexity term is fixed.

\begin{figure}[ht]
    \centering
    \includegraphics[width=0.4\textwidth]{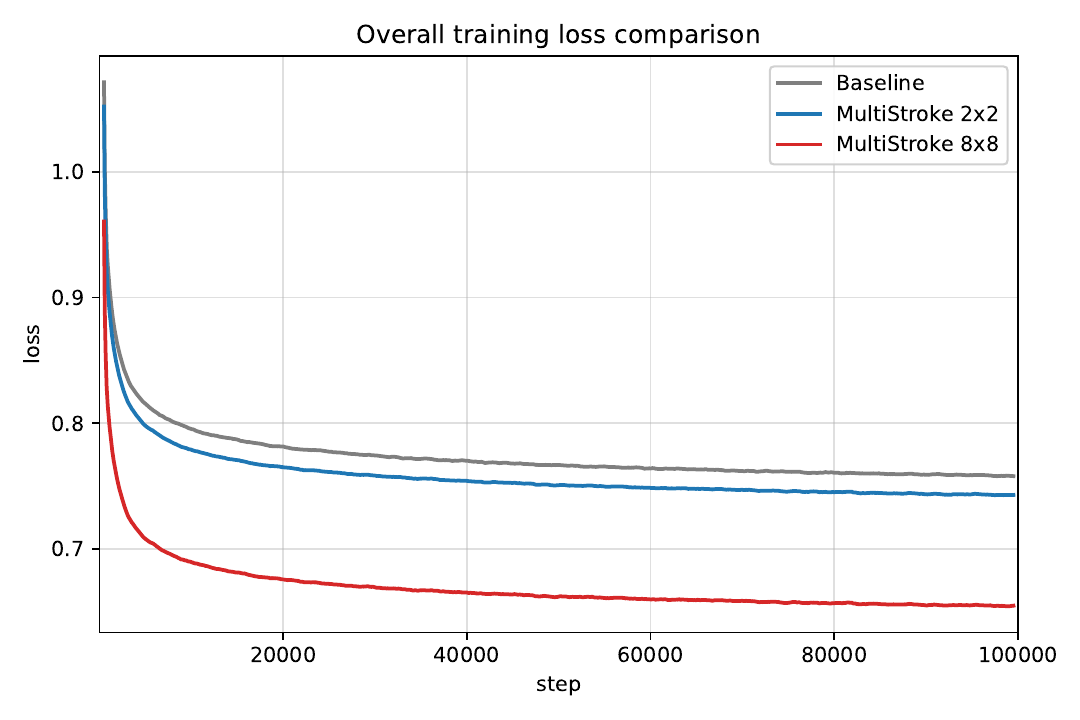}
    \caption{Training-loss comparison under matched ImageNet-1K training
    configurations. Curves show moving averages over 100 logged values from
    updates 1 to 100k.}
    \label{fig:training-loss-comparison}
\end{figure}

\paragraph{CLIPScore.}
We use the locally cached Hugging Face
\text{openai/clip-vit-large-patch14} \text{CLIPModel} and its
\text{AutoProcessor} \citep{radford2021learning}.
Each generated image is paired with its generation-time class label and mapped
through the ImageNet class-name dictionary.
After removing a trailing comma, we use the prompt
\text{a photo of a \{class\}}.
Our reported CLIP similarity is $100$ times the cosine similarity
between normalized image and text features, averaged over a fixed matched
subset of generated outputs.

\paragraph{Precision, recall, density, and coverage.}
We compute all four metrics from the same 2048-dimensional Inception features
with the fifth-nearest-neighbor radius \citep{naeem2020reliable}. The two
methods use identical generated-image identifiers and the same fixed subset of
ImageNet validation images. Distances are Euclidean, self-distances are
excluded when estimating neighborhood radii, and neighborhood membership uses
a strict inequality.

\paragraph{MultiStroke configuration.}
MultiStroke uses 100 time levels for flow matching and $8\times8$ pixel roughening. The
roughening map is area downsampling followed by nearest-neighbor
upsampling, linearly mixed with the original image according to the flow-matching time variable.
The velocity field receives a scaled VAE posterior sample of this image
and an ordinary Gaussian endpoint. It uses the same optimizer, class
conditioning, and transformer as the baseline.

The transition corrector is trained separately for 10{,}000 updates with
adjacent level pairs. It receives the scaled posterior-mean latent at the
coarser level and the coarser and finer times, then predicts the residual to
the finer-level posterior-mean latent of the same image. The corrector has four
input latent channels, 32 hidden channels, four residual convolutional blocks,
and group normalization. It is optimized with AdamW at a constant learning
rate of $10^{-4}$, weight decay $0.01$, and a global batch size of 256 across
eight GPUs, with bfloat16 forward passes and a float32 MSE loss. At each
nonfinal sampling step, the velocity field provides a clean-endpoint
estimate in the current roughness coordinates. The corrector receives this
estimate and predicts the residual to the next finer-level endpoint, which is
applied after the Euler update as specified in
Algorithm~\ref{app:algorithm-multistroke-flow}.

\begin{figure}[t]
\centering
\begin{minipage}{0.90\textwidth}
\small
\input{tables/pseudocode_flow}
\end{minipage}
\end{figure}

\input{tables/algorithm_flow_detailed}

\clearpage
\section{CIFAR-DDPM Exact Formulation and Implementation}
\label{app:cifar-ddpm}

This section specifies the CIFAR-10 \citep{krizhevsky2009learning} DDPM
experiment summarized in
Section~\ref{sec:experiments}. Algorithm~\ref{alg:multistroke-cifar-ddpm}
summarizes training and sampling; the subsequent equations specify the exact
operators, posterior updates, and clipping order.

\begin{figure}[h]
\centering
\begin{minipage}{\columnwidth}
\small
\input{tables/pseudocode_cifar_ddpm}
\end{minipage}
\end{figure}

Let $P_2$ average each non-overlapping $2\times2$ block independently in
each channel and replicate the block mean to its four pixels. For DDPM index
$t\in\{0,\ldots,N-1\}$, define the one-based level $s=t+1$ and
\begin{equation}
    w_s=\frac{\max(s-1,1)}{N},
    \qquad
    A_s=(1-w_s)I+w_sP_2.
    \label{eq:cifar-ddpm-roughening}
\end{equation}
Since $P_2$ is a projection, the roughening map and its exact inverse are
\begin{equation}
    A_s=P_2+(1-w_s)(I-P_2),
    \qquad
    A_s^{-1}=P_2+\frac{1}{1-w_s}(I-P_2).
    \label{eq:cifar-ddpm-inverse}
\end{equation}
At the coarsest level, $w_{100}=0.99$, so $A_{100}$ has condition number
$100$ and its inverse amplifies errors in the within-block residual subspace
by the same factor. The algebraic inverse is exact, but this sensitivity can
affect the coordinate transition after each posterior update. By contrast,
the adjacent transition $A_{s-1}A_s^{-1}$ amplifies that subspace by at most
a factor of two under this schedule. A lower maximum roughness or a
regularized inverse could provide greater numerical robustness.

\paragraph{Training.}
Let $\beta_t$ be the DDPM variance schedule,
$\alpha_t=1-\beta_t$, and
$\bar\alpha_t=\prod_{k=0}^{t}\alpha_k$. For clean image $x_0$ and
$\epsilon\sim\mathcal N(0,I)$, the ordinary DDPM state is
\begin{equation}
    x_t=\sqrt{\bar\alpha_t}x_0+
    \sqrt{1-\bar\alpha_t}\epsilon.
    \label{eq:cifar-ddpm-forward}
\end{equation}
MultiStroke presents $\widetilde x_t=A_sx_t$ to the network and uses
$\widetilde\epsilon_t=A_s\epsilon$ as its regression target. Its objective is
\begin{equation}
    \mathcal L_{\mathrm{MS}}(\theta)=
    \mathbb E\left[
    \operatorname{MSE}\left(
    \epsilon_\theta(\widetilde x_t,t,c),
    \widetilde\epsilon_t
    \right)\right].
    \label{eq:cifar-ddpm-loss}
\end{equation}
Linearity implies that the clean-image estimate reconstructed from these two
roughened quantities is expressed in the same $A_s$ coordinate system.

\paragraph{Sampling.}
Let the selected reverse grid contain original DDPM indices
$t_1>\cdots>t_K=0$, and let $s_i=t_i+1$. The initial Gaussian state is mapped
with $A_{s_1}$. At level $t_i$, the model predicts roughened noise and gives
the current-coordinate clean estimate
\begin{equation}
    \widehat x_{0,s_i}=
    \frac{\widetilde x_{t_i}-
    \sqrt{1-\bar\alpha_{t_i}}\,
    \epsilon_\theta(\widetilde x_{t_i},t_i,c)}
    {\sqrt{\bar\alpha_{t_i}}}.
    \label{eq:cifar-ddpm-clean-estimate}
\end{equation}
Clipping is performed directly in the current roughened coordinate system:
\begin{equation}
    \widehat x_{0,s_i}^{\mathrm{clip}}=
    \operatorname{clip}\left(\widehat x_{0,s_i},-1,1\right).
    \label{eq:cifar-ddpm-clipping}
\end{equation}
For adjacent training indices, adopt the standard boundary convention
$\bar\alpha_{-1}:=1$ and define
\begin{equation}
\begin{aligned}
    \widetilde\beta_t
    &=\beta_t\frac{1-\bar\alpha_{t-1}}{1-\bar\alpha_t},\\
    \widetilde\mu_t
    &=\frac{\sqrt{\bar\alpha_{t-1}}\beta_t}
    {1-\bar\alpha_t}\widehat x_{0,s}^{\mathrm{clip}}
    +\frac{\sqrt{\alpha_t}(1-\bar\alpha_{t-1})}
    {1-\bar\alpha_t}\widetilde x_t.
\end{aligned}
\label{eq:cifar-ddpm-posterior}
\end{equation}
With $z\sim\mathcal N(0,I)$, the stochastic update and coordinate transition
are
\begin{equation}
    y_{t-1}=\widetilde\mu_t+\sqrt{\widetilde\beta_t}A_s z,
    \qquad
    \widetilde x_{t-1}=A_{s-1}A_s^{-1}y_{t-1}.
    \label{eq:cifar-ddpm-transition}
\end{equation}
For a respaced grid, the same update uses the posterior coefficients induced
by the selected original DDPM indices and replaces $A_{s-1}$ with the
roughening map at the next selected index. At the final step, the boundary
convention gives $\widetilde\beta_0=0$ and
$y_{-1}=\widehat x_{0,1}^{\mathrm{clip}}$. The update then applies $A_1^{-1}$
and does not re-roughen the output.

\paragraph{Implementation details.}
Both class-conditional models operate directly on $32\times32$ RGB images
and use the same 38.31-million-parameter U-Net \citep{ronneberger2015unet}.
They are trained from scratch for 100{,}000 updates on two GPUs with 40 images
per GPU, Adam \citep{kingma2015adam} with learning
rate $2\times10^{-4}$, no weight decay, unit gradient clipping, and no
dropout or exponential moving average. The diffusion process uses $N=100$
steps and a linear variance schedule from $10^{-4}$ to $0.189$. Evaluation
uses all 100 reverse levels, fixed-small posterior variance, clipping of the
predicted clean image in the current roughened coordinate system, the same
uniform class sampling distribution and Gaussian prior, and the standard
StyleGAN Inception feature extractor. Table~\ref{tab:cifar-ddpm} reports the
resulting comparison.

\begin{figure}[t]
\centering
\begin{tikzpicture}
\node[anchor=south west,inner sep=0] (cifarqual) at (0,0) {%
\includegraphics[width=\textwidth]{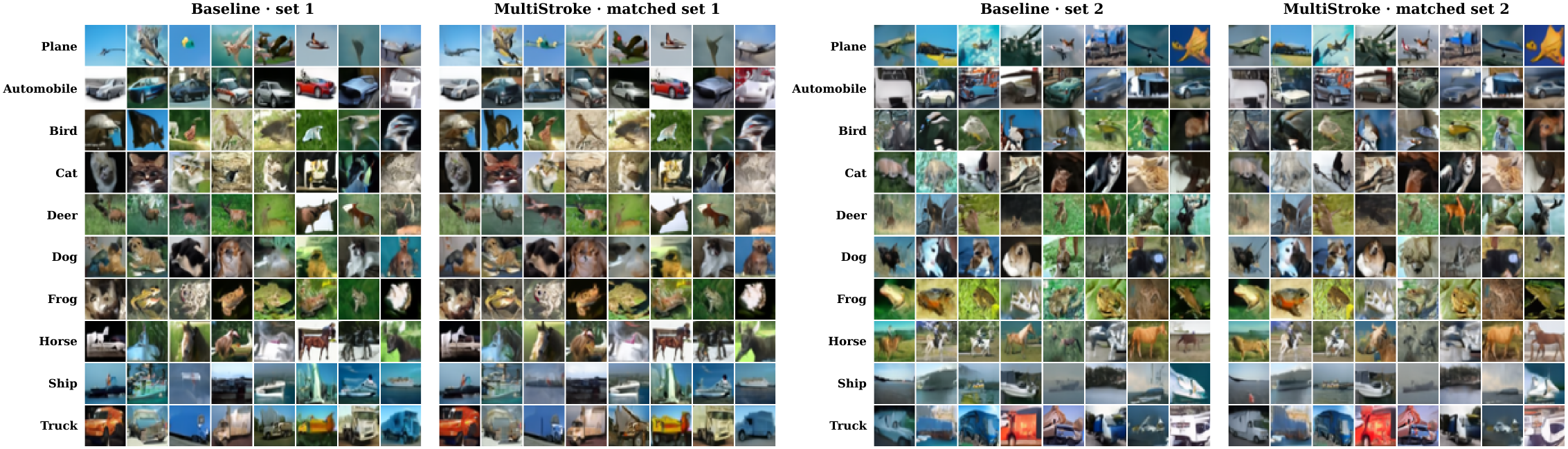}};
\begin{scope}[x={(cifarqual.south east)},y={(cifarqual.north west)}]
\draw[red,line width=1.5pt] (0.2805,0.853) rectangle (0.3065,0.944);
\draw[red,line width=1.5pt] (0.3345,0.853) rectangle (0.3605,0.944);
\draw[red,line width=1.5pt] (0.4425,0.853) rectangle (0.4685,0.944);
\draw[red,line width=1.5pt] (0.3345,0.101) rectangle (0.3605,0.191);
\end{scope}
\end{tikzpicture}
\caption{Qualitative CIFAR-10 comparison after 100k training updates. Each
MultiStroke panel uses the same class labels and initial noise as the adjacent
baseline panel; rows correspond to the CIFAR-10 classes.}
\label{fig:cifar-ddpm-qualitative}
\end{figure}

\clearpage
\section{Additional Roughening Illustration}
\begin{figure}[t]
    \centering
    \includegraphics[width=\linewidth]{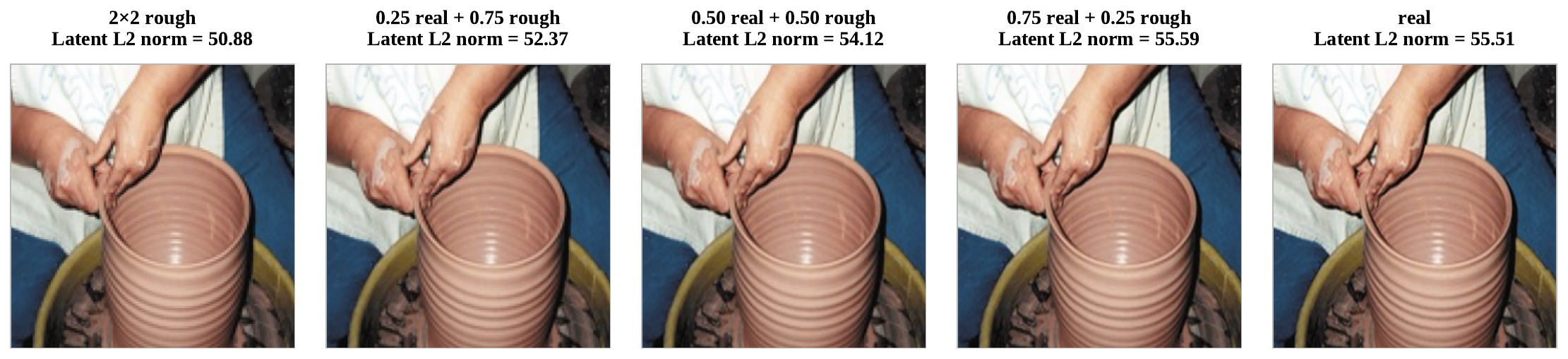}
    \caption{A $2\times2$ roughening example and interpolants, with their VAE latent
    $\ell_2$ norms. The roughened endpoint has a lower norm than the original
    image. Intermediate latent norms need not decrease monotonically because
    the nonlinear VAE encoder does not preserve pixel-space norm ordering.}
    \label{fig:latent-norm-2x2}
\end{figure}

Figure~\ref{fig:latent-norm-2x2} gives a representative $2\times2$
pixel-roughening interpolation and the corresponding VAE latent norms.
It visualizes the attenuation of within-block detail described in
Section~\ref{sec:theory}.

\input{sections/theory_app.tex}

%% file: sections/2_related_extended.tex
\section{Extended Related Work}
\label{app:related}

Diffusion models learn to reverse a progressive noising process
\citep{ho2020ddpm}, while score-based models describe generation through
continuous-time stochastic dynamics \citep{song2021score}. Flow matching
instead learns a time-dependent vector field for a prescribed probability
path \citep{lipman2023flow}. Closely related formulations include rectified
flow, which promotes straighter transport trajectories
\citep{liu2023rectified}, and stochastic interpolants, which unify
deterministic flows and stochastic diffusion processes
\citep{albergo2023stochastic}. Latent generative models reduce computation by
moving these dynamics into the representation space of a pretrained
autoencoder \citep{rombach2022ldm,vahdat2021lsgm}. MultiStroke adopts latent
flow matching but changes the spatial granularity of both the interpolated
state and its supervised endpoint before encoding.

A complementary line of work replaces isotropic Gaussian noising with
structured corruptions. Inverse heat dissipation and progressive deblurring
construct coarse-to-fine generative processes in the frequency domain
\citep{rissanen2023heat,lee2022progressive}. Blurring Diffusion progressively
suppresses high-frequency content \citep{hoogeboom2023blurring}, Cold
Diffusion generalizes iterative restoration to deterministic image
degradations \citep{bansal2023cold}, and Soft Diffusion develops score
matching for general linear corruptions \citep{daras2023softdiffusion}.
Iterative alpha-deblending obtains a deterministic generative process through
repeated blending and inversion \citep{heitz2023iterative}. These studies show
that the corruption path can encode useful image structure. MultiStroke does
not introduce an alternative diffusion process; it places a time-dependent
block-averaging transformation inside a latent flow-matching path and applies
it jointly to the input and regression target.

Coarse-to-fine generation has also been implemented through explicit
resolution hierarchies. Cascaded diffusion models use separate generative and
super-resolution stages \citep{ho2022cascaded}, while f-DM applies progressive
signal transformations across multiple stages \citep{gu2023fdm}. Simple
Diffusion uses a multiresolution objective to support high-resolution
generation within a unified model \citep{hoogeboom2023simple}. In contrast,
MultiStroke retains a single image resolution and a single velocity model,
varying target granularity along the path. Its learned corrector addresses the
nonlinear VAE transition between adjacent roughness levels rather than serving
as a separate super-resolution generator.

MultiStroke further relates to multiscale representation learning and
regularization. U-Net-style processing changes internal feature resolution
\citep{ronneberger2015unet}, curriculum learning orders examples by difficulty
\citep{bengio2009curriculum}, and dropout regularizes model parameters
\citep{srivastava2014dropout}. MultiStroke instead controls the spatial scale
of the regression problem at each path time. This motivation is consistent
with the low-frequency learning bias of neural networks
\citep{rahaman2019spectral} and with persistent spectral discrepancies in
generated images \citep{dzanic2019fourier}, while remaining distinct from an
architecture-level multiscale design or a training-stage curriculum.

%% file: tables/pseudocode_flow.tex
\captionsetup{type=algorithm}
\captionof{algorithm}{Pseudocode: MultiStroke for Flow Matching}
\label{alg:multistroke-flow-corrector}
\begin{algorithmic}[1]
\INPUT{image $x_0$, class label $c$, steps $N$, pixel roughening map $R$}
\STATE \textbf{Train the velocity field:} sample time $t$ and noise $\epsilon$
\STATE Roughen $x_0$ at time $t$, encode it with the frozen VAE, and form the ordinary flow-matching interpolation with $\epsilon$
\STATE Regress the velocity model to the difference between the Gaussian endpoint and the encoded roughened image
\STATE \textbf{Train the corrector:} sample adjacent levels $t$ and $t-h$
\STATE Encode the two correspondingly roughened images with VAE posterior means
\STATE Regress the corrector from the coarser-level latent and the two levels to the residual toward the finer-level latent
\STATE \textbf{Sample:} initialize $x$ from a standard Gaussian latent
\FOR{each Euler level $t$ from noise to data}
    \STATE Predict the velocity $v$
    \STATE Estimate the clean endpoint in the current roughness coordinates
    \STATE Apply the Euler update $x\gets x-hv$
    \IF{a finer level remains}
        \STATE Feed the current-level endpoint estimate and the two levels to the corrector
        \STATE Predict the residual to the next finer-level endpoint
        \STATE Add the residual scaled by the remaining clean-endpoint weight
    \ENDIF
\ENDFOR
\STATE \textbf{return} $\operatorname{VAEDec}(x)$
\end{algorithmic}

%% file: tables/algorithm_flow_detailed.tex
\begin{figure}[t]
\centering
\begin{minipage}{0.90\textwidth}
\small
\captionsetup{type=algorithm}
\captionof{algorithm}{Complete MultiStroke Training and Sampling Algorithm}
\label{app:algorithm-multistroke-flow}
\begin{algorithmic}[1]
\INPUT{$x_0,c$, flow-matching steps $N\geq2$, $h\gets1/N$, block projection $P_b$, frozen VAE encoder $\operatorname{Enc}$ and decoder $\operatorname{Dec}$}
\STATE $R_t(x)\gets(1-t+h)x+(t-h)P_bx$ for $t\in[h,1]$
\STATE \textbf{Velocity training:} sample $u\sim\operatorname{Unif}[0,1]$, set $t\gets\max(u,h)$, and sample $\epsilon\sim\mathcal{N}(0,I)$
\STATE $z_t\gets0.18215\operatorname{Enc}(R_t(x_0))$; $x_t\gets(1-t)z_t+t\epsilon$
\STATE $\mathcal{L}_v\gets\operatorname{MSE}(v_\theta(x_t,t,c),\epsilon-z_t)$
\STATE \textbf{Corrector training:} sample $t\sim\operatorname{Unif}\{2h,3h,\ldots,1\}$
\STATE $z_t^\mu\gets0.18215\operatorname{Enc}_\mu(R_t(x_0))$; $z_{t-h}^\mu\gets0.18215\operatorname{Enc}_\mu(R_{t-h}(x_0))$
\STATE $\mathcal{L}_c\gets\operatorname{MSE}(c_\phi(z_t^\mu,t,t-h),z_{t-h}^\mu-z_t^\mu)$
\STATE \textbf{Sampling:} $x\gets\eta$, where $\eta\sim\mathcal{N}(0,I)$
\FOR{$j=N,N-1,\ldots,1$}
    \STATE $t\gets jh$; $t'\gets(j-1)h$; $v\gets(1+w)v_\theta(x,t,c)-w v_\theta(x,t,\varnothing)$
    \STATE $\widehat{z}_t\gets x-tv$; $x\gets x-hv$
    \IF{$j>1$}
        \STATE $x\gets x+(1-t')c_\phi(\widehat{z}_t,t,t')$
    \ENDIF
\ENDFOR
\STATE \textbf{return} $\operatorname{Dec}(x/0.18215)$
\end{algorithmic}
\end{minipage}
\end{figure}

%% file: tables/pseudocode_cifar_ddpm.tex
\captionsetup{type=algorithm}
\captionof{algorithm}{MultiStroke training and sampling for CIFAR-DDPM.}
\label{alg:multistroke-cifar-ddpm}
\begin{algorithmic}[1]
\INPUT{clean image, class label, DDPM schedule, block-averaging operator}
\STATE Training: sample a diffusion level and Gaussian noise
\STATE Form the ordinary noisy DDPM image
\STATE Roughen both the noisy image and Gaussian-noise target at that level
\STATE Regress the noise-prediction network from the roughened image to the roughened target
\STATE Sampling: roughen the initial Gaussian image at the first selected level
\FOR{each selected level from noise to data}
    \STATE Predict noise in the current roughened coordinates
    \STATE Reconstruct the predicted clean image in those coordinates
    \STATE Clip the predicted clean image in the current roughened coordinates
    \STATE Form the stochastic DDPM posterior update at the current level
    \STATE Invert the current roughening and apply the next selected level's roughening
\ENDFOR
\STATE Return the final inverse-roughened image
\end{algorithmic}

%% file: sections/theory_app.tex
\clearpage
\section{Supplementary Theoretical Results and Proofs}
\subsection{Proof of Lemma~\ref{lem:mse_rad}}
\label{app:regression-proof}
\begin{proof}
By standard symmetrization \citep{bartlett2002rademacher},
$\mathbb{E}_S[\sup_{h\in\mathcal{H}}(R(h)-\widehat{R}_S(h))]
\leq2\mathfrak{R}_n(\mathcal{L})$.
Since $\widehat{h}\in\mathcal{H}$, this also bounds its expected generalization gap.

Fix $S$. For sample $i$, define
$\phi_i(u)=\sum_{p=1}^d(u_p-y_{ip})^2-\sum_{p=1}^d y_{ip}^2$.
Subtracting the target-only term leaves the empirical Rademacher
complexity unchanged because its signed sum has expectation zero. For
$u,v\in[-B,B]^d$, the line segment joining them remains in this cube, and
the gradient obeys
\[
\|\nabla\phi_i(u)\|_2
=2\|u-y_i\|_2
\leq2\sqrt d(B+Y).
\]
Thus $\phi_i$ is $2\sqrt d(B+Y)$-Lipschitz in the Euclidean norm on the
prediction range. Applying the vector-contraction inequality
\citep{maurer2016vector} gives
\begin{equation}
\mathfrak{R}_n(\mathcal{L})
\leq
2\sqrt{2d}(B+Y)
\mathfrak R_n^{\mathrm{vec}}(\mathcal H).
\label{eq:mse_contraction}
\end{equation}
Combining the contraction bound with symmetrization proves Eq.~\eqref{eq:mse_gen_bound}.
\end{proof}

\subsection{Proof of Theorem~\ref{prop:independent-pixel-benchmark}}
\label{app:block-average-proof}

\begin{proof}
Fix a sample $S$ with pairwise distinct inputs $x_1,\ldots,x_n$.
We first compute the empirical complexity of the unrestricted class
$\mathcal H_\square$, and then repeat the calculation after block
averaging. The distinctness assumption ensures that, for this class,
every coordinate $h_p(x_i)$ can be chosen independently within
$[-B,B]$.

For a fixed collection of Rademacher signs,
$\sup_{a\in[-B,B]}\sigma_{ip}a=B|\sigma_{ip}|=B$.
Consequently,
\begin{align*}
&\sup_{h\in\mathcal H_\square}
\sum_{i=1}^n\sum_{p=1}^d
\sigma_{ip}h_p(x_i)\\
&\quad=\sum_{i=1}^n\sum_{p=1}^d
\sup_{a\in[-B,B]}\sigma_{ip}a\\
&\quad=B\sum_{i=1}^n\sum_{p=1}^d|\sigma_{ip}|
=Bnd.
\end{align*}
Taking the expectation over the signs and dividing by $n$ gives
\begin{equation}
\widehat{\mathfrak R}^{\mathrm{vec}}_S(\mathcal H_\square)
=\frac1n\mathbb E_\sigma[Bnd]=Bd.
\label{eq:benchmark-unroughened-proof}
\end{equation}
Thus, before roughening, the unrestricted predictor can align
independently with the sign of every output coordinate.

For general $\alpha$, symmetry of $T_\alpha$ and coordinatewise
maximization give
\begin{align*}
\widehat{\mathfrak R}^{\mathrm{vec}}_S
(T_\alpha\mathcal H_\square)
&=\frac Bn\sum_{i=1}^n
\mathbb E_\sigma\|T_\alpha\boldsymbol\sigma_i\|_1,\\
\frac1d\widehat{\mathfrak R}^{\mathrm{vec}}_S
(T_\alpha\mathcal H_\square)
&=B\mathbb E_\sigma\left|
(1-\alpha)\sigma_1+\frac{\alpha}{m}
\sum_{j=1}^m\sigma_j\right|.
\end{align*}
For
$Z=(1-\alpha)\sigma_1+\alpha m^{-1}\sum_{j=1}^m\sigma_j$,
independence of the signs gives
\[
\mathbb E_\sigma Z^2
=(1-\alpha)^2+\frac{2\alpha-\alpha^2}{m}
=1-(2\alpha-\alpha^2)\left(1-\frac1m\right).
\]
Cauchy-Schwarz gives $\mathbb E_\sigma|Z|\leq
(\mathbb E_\sigma Z^2)^{1/2}$, proving
Eq.~\eqref{eq:independent_pixel_bound}.

Next consider the class $A\mathcal H_\square$.
Since the block-averaging operator is symmetric ($A^\top=A$),
we may transfer it from the prediction to the random signs:
\begin{align*}
&\widehat{\mathfrak R}^{\mathrm{vec}}_S
(A\mathcal H_\square)\\
&\quad=\frac1n\mathbb E_\sigma
\sup_{h\in\mathcal H_\square}
\sum_{i=1}^n\boldsymbol\sigma_i^\top Ah(x_i)\\
&\quad=\frac1n\mathbb E_\sigma
\sup_{h\in\mathcal H_\square}
\sum_{i=1}^n(A\boldsymbol\sigma_i)^\top h(x_i)\\
&\quad=\frac Bn\sum_{i=1}^n
\mathbb E_\sigma\|A\boldsymbol\sigma_i\|_1.
\end{align*}
The last equality again follows by maximizing each coordinate
independently over $[-B,B]$.

Let $\mathcal B$ denote one block of $m$ output coordinates.
For every $p\in\mathcal B$, block averaging gives
\[
(A\boldsymbol\sigma_i)_p
=\frac1m\sum_{q\in\mathcal B}\sigma_{iq}.
\]
All $m$ coordinates in this block consequently have the same
averaged value. Their total contribution to the $\ell_1$ norm is
\begin{align*}
\sum_{p\in\mathcal B}
|(A\boldsymbol\sigma_i)_p|
&=m\left|\frac1m
\sum_{q\in\mathcal B}\sigma_{iq}\right|\\
&=\left|\sum_{q\in\mathcal B}\sigma_{iq}\right|.
\end{align*}
Because the $d$ coordinates are partitioned into $d/m$ blocks,
and the independent signs have the same distribution in every block,
linearity of expectation yields
\begin{align*}
\mathbb E_\sigma\|A\boldsymbol\sigma_i\|_1
&=\sum_{\mathcal B}\mathbb E_\sigma
\left|\sum_{q\in\mathcal B}\sigma_{iq}\right|\\
&=\frac dm\mathbb E_\sigma
\left|\sum_{j=1}^m\sigma_j\right|.
\end{align*}
Substituting this expression and using the identical distribution
across the $n$ samples, we obtain
\begin{equation}
\widehat{\mathfrak R}^{\mathrm{vec}}_S
(A\mathcal H_\square)
=\frac{Bd}{m}\mathbb E_\sigma
\left|\sum_{j=1}^m\sigma_j\right|.
\label{eq:benchmark-roughened-proof}
\end{equation}
Dividing Eq.~\eqref{eq:benchmark-roughened-proof} by
Eq.~\eqref{eq:benchmark-unroughened-proof} establishes the
exact ratio in Eq.~\eqref{eq:independent_pixel_ratio}.
The difference between the two cases is that, after averaging,
one shared block value must align with the \emph{sum} of its $m$
independent signs rather than with each sign separately.

We next derive the stated upper bound. By Cauchy-Schwarz,
\begin{equation}
\mathbb E_\sigma\left|\sum_{j=1}^m\sigma_j\right|
\leq\left[\mathbb E_\sigma
\left(\sum_{j=1}^m\sigma_j\right)^2\right]^{1/2}.
\label{eq:benchmark-cauchy-proof}
\end{equation}
Expanding the square and using independence,
$\mathbb E\sigma_j=0$, and $\mathbb E\sigma_j^2=1$, gives
\begin{align*}
\mathbb E_\sigma
\left(\sum_{j=1}^m\sigma_j\right)^2
&=\sum_{j=1}^m\mathbb E\sigma_j^2
+2\sum_{j<k}\mathbb E(\sigma_j\sigma_k)\\
&=m+2\sum_{j<k}
\mathbb E\sigma_j\,\mathbb E\sigma_k\\
&=m.
\end{align*}
Combining this identity with
Eq.~\eqref{eq:benchmark-cauchy-proof} gives
\[
\mathbb E_\sigma\left|\sum_{j=1}^m\sigma_j\right|
\leq\sqrt m.
\]
Equation~\eqref{eq:benchmark-roughened-proof} therefore implies
$\widehat{\mathfrak R}^{\mathrm{vec}}_S
(A\mathcal H_\square)/d\leq B/\sqrt m$. Together with
Eq.~\eqref{eq:benchmark-unroughened-proof}, this gives the upper bound in
Eq.~\eqref{eq:independent_pixel_ratio}.

We finally compute the expectation exactly when $m$ is even.
Let $K$ count the number of negative signs among
$\sigma_1,\ldots,\sigma_m$. There are $\binom{m}{k}$ ways to select exactly $k$ negative
signs, and each of the $2^m$ sign configurations has probability
$2^{-m}$. Therefore,
\[
K\sim\operatorname{Binomial}(m,1/2),\qquad
\mathbb P(K=k)=\binom{m}{k}2^{-m}.
\]
Conditional on $K=k$, there are $m-k$ positive signs and
$k$ negative signs. Thus their sum is
\begin{equation}
\sum_{j=1}^m\sigma_j
=(m-k)-k=m-2k,
\qquad\text{when }K=k.
\label{eq:benchmark-signcount-proof}
\end{equation}
Using the binomial probabilities to evaluate the expectation,
\begin{equation}
\begin{aligned}
\mathbb E_\sigma
\left|\sum_{j=1}^m\sigma_j\right|
&=\mathbb E|m-2K|\\
&=\frac1{2^m}\sum_{k=0}^m
|m-2k|\binom{m}{k}.
\end{aligned}
\label{eq:benchmark-binomial-expectation-proof}
\end{equation}
Write $m=2r$, where $r=m/2$ is an integer.
The term $k=r$ vanishes because $m-2r=0$. Moreover,
for every $k<r$, the term indexed by $m-k$ has the same
magnitude and binomial coefficient as the term indexed by $k$:
\[
\binom{m}{m-k}=\binom{m}{k},\qquad
|m-2(m-k)|=m-2k.
\]
We can therefore pair the terms below and above $r$ in
Eq.~\eqref{eq:benchmark-binomial-expectation-proof}, obtaining
\begin{equation}
\mathbb E|m-2K|
=\frac2{2^m}\sum_{k=0}^{r-1}
(m-2k)\binom{m}{k}.
\label{eq:benchmark-symmetric-sum-proof}
\end{equation}

To evaluate the remaining finite sum, we first express the
binomial coefficients at level $m-1$ in terms of those at level
$m$. By the factorial definition, for $0\leq k\leq r-1$,
\begin{align*}
\binom{m-1}{k}
&=\frac{(m-1)!}{k!(m-1-k)!}\\
&=\frac{m-k}{m}\binom{m}{k}.
\end{align*}
Similarly, for $1\leq k\leq r-1$,
\begin{align*}
\binom{m-1}{k-1}
&=\frac{(m-1)!}{(k-1)!(m-k)!}\\
&=\frac{k}{m}\binom{m}{k}.
\end{align*}
The second identity also holds at $k=0$ under the convention
$\binom{m-1}{-1}=0$.
Subtracting the second identity from the first and multiplying
by $m$ gives
\begin{equation}
(m-2k)\binom{m}{k}
=m\left[\binom{m-1}{k}-\binom{m-1}{k-1}\right].
\label{eq:benchmark-telescoping-identity-proof}
\end{equation}
Summing
Eq.~\eqref{eq:benchmark-telescoping-identity-proof}
from $k=0$ to $r-1$ yields
\begin{align*}
&\sum_{k=0}^{r-1}(m-2k)\binom{m}{k}\\
&=m\left[\sum_{k=0}^{r-1}\binom{m-1}{k}
-\sum_{k=0}^{r-1}\binom{m-1}{k-1}\right]\\
&=m\left[\sum_{k=0}^{r-1}\binom{m-1}{k}
-\sum_{j=-1}^{r-2}\binom{m-1}{j}\right]\\
&=m\binom{m-1}{r-1}.
\end{align*}
The penultimate line merely reindexes the second sum
with $j=k-1$; the last line follows because all common
terms cancel. Finally, the remaining coefficient can be expressed in terms
of the central binomial coefficient:
\begin{equation}
\begin{aligned}
\binom{m-1}{r-1}
&=\frac{(m-1)!}{(r-1)!(m-r)!}\\
&=\frac{r}{m}\frac{m!}{r!(m-r)!}\\
&=\frac{r}{m}\binom{m}{r}\\
&=\frac12\binom{m}{m/2},
\end{aligned}
\label{eq:benchmark-central-binomial-proof}
\end{equation}
Substituting these last two identities into
Eq.~\eqref{eq:benchmark-symmetric-sum-proof} gives
\begin{align*}
\mathbb E_\sigma\left|\sum_{j=1}^m\sigma_j\right|
&=\frac{2m}{2^m}\binom{m-1}{r-1}\\
&=\frac{m}{2^m}\binom{m}{m/2}.
\end{align*}
Dividing by $m$ therefore yields the exact normalized
complexity ratio
\[
\frac1m\mathbb E_\sigma\left|\sum_{j=1}^m\sigma_j\right|
=2^{-m}\binom{m}{m/2}.
\]

For illustration, when $m=4$, the binomial counts for
$K=0,1,2,3,4$ are $1,4,6,4,1$, respectively. Hence
\begin{align*}
\mathbb E|4-2K|
&=\frac{4(1)+2(4)+0(6)+2(4)+4(1)}{16}\\
&=\frac32,
\end{align*}
which gives the $2\times2$ ratio $(3/2)/4=3/8$.
For an $8\times8$ block, $m=64$ and the same formula gives
$2^{-64}\binom{64}{32}\approx0.0993$.
\end{proof}